\documentclass[runningheads]{llncs}
\usepackage[T1]{fontenc}
\usepackage{graphicx}
\usepackage{lipsum} 
\usepackage{hyperref}
\usepackage{url}

\usepackage{tikz}
\def\checkmark{\tikz\fill[scale=0.25](0,.35) -- (.25,0) -- (1,.7) -- (.25,.15) -- cycle;}

\begin{document}
\title{Infused Suppression Of Magnification Artefacts For Micro-AU Detection}
% \name{Huai-Qian Khor$^{1}$, Yante Li$^{1}$, Xingxun Jiang$^{2,1}$$^{}$ , Guoying Zhao$^{1\star}$
% \thanks{*Corresponding Author E-mail: guoying.zhao@oulu.fi}
% }
% \address{$^{1}$ University of Oulu, $^{2}$ Southeast University}

\author{Huai-Qian Khor\inst{1} \and
Yante Li\inst{1} \and
Xingxun Jiang\inst{2} \and Guoying Zhao\inst{1}$^{\star}$ }

% \footnote{}

% {firstname.lastname}@oulu.fi
\institute{Center for Machine Vision and Signal Analysis, University of Oulu, Finland \email{huai.khor@oulu.fi, yante.li@oulu.fi} \and
School of Biological Sciences and Medical Engineering, Southeast University
\email{jiangxingxun@seu.edu.cn}\\
$^\star$Corresponding Author, E-mail: guoying.zhao@oulu.fi
}
\maketitle              % typeset the header of the contribution
\addtocounter{footnote}{-2}
\footnotetext{Code is publicly available at \url{ https://github.com/IcedDoggie/InfuseNet_SCIA2025}}
\begin{abstract}
Facial micro-expressions are spontaneous, brief and subtle facial motions that unveil the underlying, suppressed emotions. Detecting Action Units (AUs) in micro-expressions is crucial because it yields a finer representation of facial motions than categorical emotions, effectively resolving the ambiguity among different expressions. One of the difficulties in micro-expression analysis is that facial motions are subtle and brief, thereby increasing the difficulty in correlating facial motion features to AU occurrence. To bridge the subtlety issue, flow-related features and motion magnification are a few common approaches as they can yield descriptive motion changes and increased motion amplitude respectively. While motion magnification can amplify the motion changes, it also accounts for illumination changes and projection errors during the amplification process, thereby creating motion artefacts that confuse the model to learn inauthentic magnified motion features. The problem is further aggravated in the context of a more complicated task where more AU classes are analyzed in cross-database settings.

To address this issue, we propose \textbf{InfuseNet}, a layer-wise unitary feature infusion framework that leverages motion context to constrain the Action Unit (AU) learning within an informative facial movement region, thereby alleviating the influence of magnification artefacts. On top of that, we propose leveraging magnified latent features instead of reconstructing magnified samples to limit the distortion and artefacts caused by the projection inaccuracy in the motion reconstruction process. Via alleviating the magnification artefacts, \textbf{InfuseNet} has surpassed the state-of-the-art results in the CD6ME protocol. Further quantitative studies have also demonstrated the efficacy of motion artefacts alleviation.
\end{abstract}
\begin{keywords}
AU Detection, Magnification Artefacts, Infusion
\end{keywords}

\section{Introduction}
\label{sec:intro}
% elaboration: 2 more paragraphs
% Background paragraph % 
% write about what makes a micro-expression micro, how they occur, what is AU and why we study AU %
% Talks about limited data problem %
Micro-expressions (MEs) are spontaneous, subtle and rapid (1/25 to 1/3 seconds) \cite{oh2018survey} facial movements solicited under high-stakes environments \cite{ekman1971constants} \cite{ekman2009lie} such as psychological diagnosis, interviews and interrogations. MEs are elicited when the subject attempts to suppress the emotion in a stressful environment, eventually causing a minor facial tweak that reveals the underlying emotion. Compared to conventional facial expressions, MEs' subtlety and involuntariness increase the difficulty of identifying the emotion or AUs, this complicates the data collection and annotation process because it requires trained and certified Facial Action Coding System (FACS) coders to perform exhaustive and meticulous data annotation to identify the AUs and their occurrences' timeframe (the onset, apex and offset frames). As a result, this whole process causes the data collection to be slower and more expensive, leading to small datasets. 

In ME annotation, FACS encodes facial muscle movements into action units (AUs) whereby each AU has its description, and one or multiple AUs constitute an emotional category. For instance, AU6 (Cheek Raiser) and AU12 (Lip Corner Puller) translate to a positive or happy emotion. During the annotation process, the ME emotion categorization involves using a mixture of self-reports and AUs \cite{yan2014casme}. Given a scenario, where data annotators label the existence of AU6 and AU12 which translates to a happy emotion, if the self-report from the participant indicates that the emotional experience is otherwise, the AU labels will be overridden by the self-reported emotional experience instead. As a result, there is a certain level of ambiguity in the current ME datasets' emotional categorization because the definition of emotion may not correspond to the related AUs. Therefore, studying the detection of AUs is critical because it yields a finer granularity in micro-facial motion analysis than recognizing emotions, and this reduces the ambiguity in analyzing individual emotions \cite{niu2019local}.

\begin{figure}[t]
    \centering
    \includegraphics[width=\columnwidth]{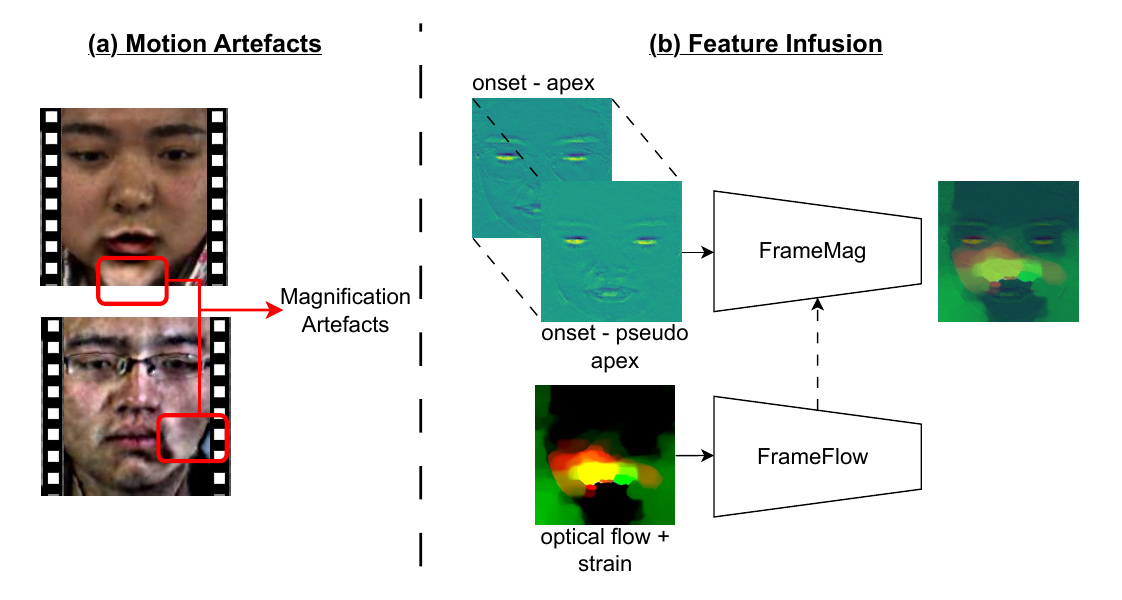}
    % \vspace{-1.5em}
    \caption{The illustrations of magnification artefacts and feature infusion. Part (A) demonstrates the magnification artefacts generated during the motion amplification, where faces appear deformed. Part (B) presents the proposed feature infusion process, which extracts latent magnified features while counterbalancing the amplified noise using optical flow images.}
    % \vspace{-2.0em}
    \label{fig:concept}
\end{figure}

% magnification here %
% to-do expands
In signal analysis, learnable Eulerian Video Magnification (EVM) \cite{oh2018learning} can magnify subtle motion changes such as breathing patterns and blood circulations. Its application is suitable for solving one of the critical challenges in analyzing MEs: subtle facial changes. While it has been explored on several occasions in ME recognition \cite{li2021intra}\cite{li2021micro_kd}\cite{li2021micro_sca}\cite{wei2023cmnet}\cite{varanka2023learnable}, motion magnification generates motion artefacts during the amplified reconstruction process as demonstrated in Figure \ref{fig:concept}. The causes of motion artefacts are the over-amplification of motion due to illumination inconsistency and inaccurate projection from linear latent features to non-linear reconstruction of magnified images. As a result, the motion artefacts can be detrimental in more complicated tasks such as fine-grained and multi-dataset AU detection tasks. This motivates us to study the alleviation of artefacts to solve a comparatively more challenging task.

% One of the critical challenges of analyzing MEs is the subtle facial changes. Motion magnification \cite{oh2018learning} has been utilized to amplify the magnitude of the motion changes. 

% optical flow here %
One of the ways to alleviate the motion artefacts is to provide soft guidance as a prior for identifying authentic facial motions, where the soft guidance signal should accurately capture and convey motion dynamics. Conveniently, optical flow is widely used to compute pixel displacement, capturing motion changes while mitigating the influence of illumination variations. To improve representativeness on tinier motions, subsequent research has leveraged the second derivative of optical flow to compute optical strain \cite{shreve2009towards}, further enhancing the encoding of subtle facial movements. Given the motion representation features of optical flow, it motivates wide adoption in facial micro-expression study \cite{cai2024mfdan} \cite{gan2019off} because it can detect tiny facial motion changes well. In conjunction, we can leverage its motion representativeness to provide a soft attention mask to steer the feature learning process of magnified features.

% proposal goes here % 
To address the motion artefact challenge, we propose InfuseNet, which alleviates motion artefacts with optical flow features in a layer-successive approach. We deduce that unitarily infusing descriptive motion features into magnified motion features enables the model to learn magnified yet descriptive motion features. Motivated by the adaptation of EVM \cite{varanka2023learnable}, we attempt to constrain the motion artefacts caused by reconstruction via removing the motion magnification decoder and mainly exploiting the latent features.

In summary, our contributions are as follows:
\begin{itemize}
    \item Designing unitary successive layer features to alleviate magnified motion artefacts for micro AU detection.
    \item Utilizing latent magnified features to minimize the motion artefacts generated from image decoding.
    \item Achieving state-of-the-art Micro AU detection results under the CD6ME protocol and qualitatively analyzing the efficacy of infusion.
    % \item Studying the problem of magnification artefacts and the effects of artefact removal on micro-AU detection with CD6ME protocol.    
\end{itemize}

\section{Related Work}
% In AU Detection, talks about what is AU detection, %
% talks about what had been done in macro %
% discusses what had been done in micro and their weaknesses %

% elaboration: 2 more paragraphs for each

In this section, we examine past works on AU detection and motion magnification to understand their shortcomings and the motivations gained from them.

\subsection{AU Detection}
AU is a discretized representation of facial muscle movement that can translate to an affective message \cite{zhi2020comprehensive} such as emotions. AU detection studies were recently explored in macro-level AU detection \cite{yuan2025auformer}. Yuan et al. \cite{yuan2025auformer} designed a transformer-based approach and Mixture-of-Knowledge Expert to study aggregated information for parameter-efficient AU detection, mainly in macro AU settings. 

In micro-expression analysis, Yin et al. \cite{yin2023aware} used an action-unit-aware graph convolutional network to model the relationship between facial ROI for the ME spotting task, whereby action unit only served as the auxiliary context to assist ME spotting. For micro AU detection, several early works \cite{li2021intra} \cite{li2021micro_kd} \cite{li2021micro_sca} had attempted to study its efficacy. Li et al. \cite{li2021micro_sca} utilized relationship information of facial regions via spatial- and channel-wise features to study micro AUs. Subsequently, Li et al. \cite{li2021micro_kd} \cite{li2021intra} leverage contexts from macro AUs via knowledge distillation and contrastive learning to translate the knowledge to micro AU detection. Despite that, the extent of the works above was constrained within a singular database analysis and lesser AU classes. Motivated by these milestones and shortcomings, we continue to build on top of AU detection but in a more challenging and realistic scenario where joint databases and larger AU classes are considered.

\subsection{Motion Magnification}
Motion magnification is prevalent in ME analysis because of its ability to augment micro motions. Its early adaptations in ME spotting and recognition trace back to \cite{park2015subtle} \cite{le2016eulerian} \cite{li2015reading}. Park et al. \cite{park2015subtle} proposed an adaptive magnification approach that selectively enhances discriminative facial regions to improve subtle expression recognition. Similarly, Le et al. \cite{le2016eulerian} incorporated magnification with spatiotemporal feature modifications, including spatial decomposition to preserve fine-grained details and bandpass filtering to isolate relevant temporal variations. However, the magnification factor remains limited, as excessive amplification may introduce noise, thereby degrading recognition performance.

As it is proven effective, newer methods have looked into utilizing a learnable EVM \cite{oh2018learning} or built a learning mechanism on top of vanilla EVM in ME analysis. Notably, Wei et al. \cite{wei2022novel} worked on adaptive magnification levels to recognize ME emotions in sequence. It designed a magnification attention module to determine the optimal magnification level for each micro-expression adaptively. Varanka et al. \cite{varanka2023learnable} proposed a learnable mechanism to magnify the motion dynamics selectively for singular dataset AU detection. It added a learnable module on top of classical EVM to craft learnable magnified motion changes without harnessing optical flow-related features. 

% why study artefacts?
Despite the creation of adaptive motion magnification, motion artefacts are still persistent. The existence of motion artefacts limits the magnification factor and it is worse when the complexity of the task increases, thereby limiting the study of micro AU in more challenging settings. Therefore, it motivates us to study this problem as we perform AU detection in six micro-expression joint database settings.

% \cite{wei2023cmnet} \cite{wei2022novel} \cite{varanka2023learnable}
 % To-do: describe how vit-tiny works and how infusion works
% % % \vspace{-1.2em}
\section{Methodology}

% To-do: can elaborate about ViT 

\begin{figure*}[!h]
    \centering
    \includegraphics[width=1.0\linewidth]{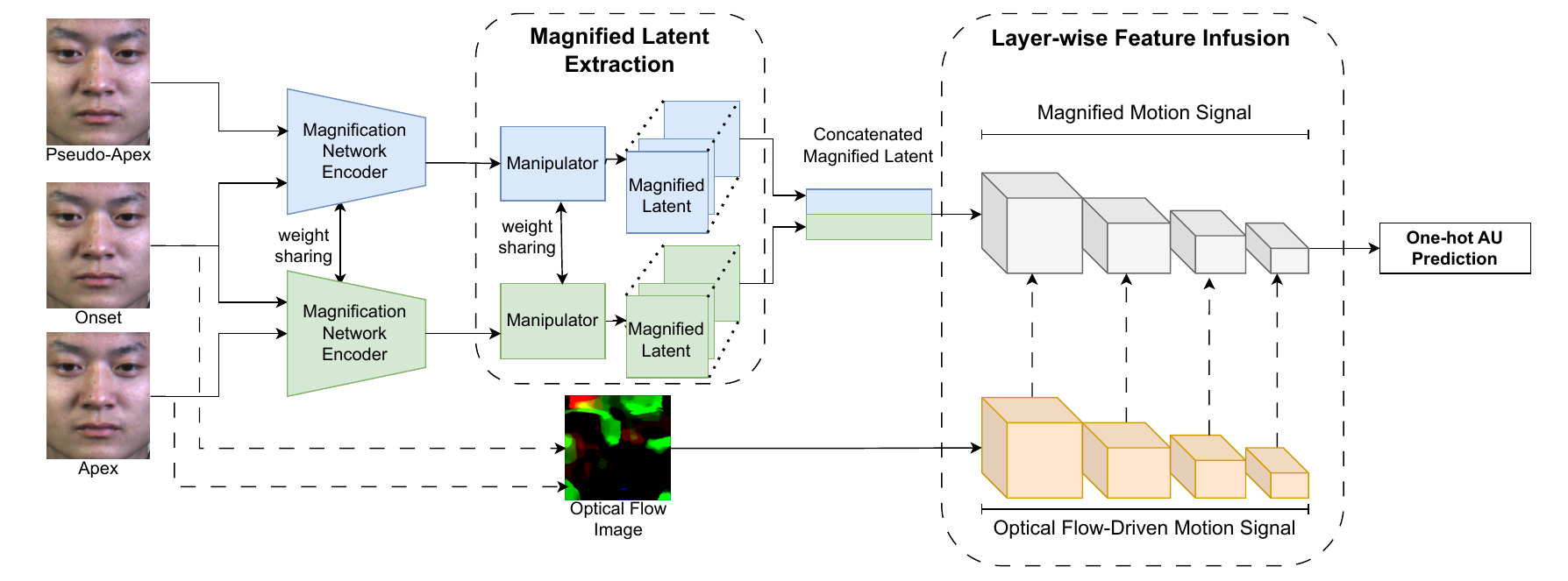}
    % \vspace{-1.5em}

    \caption{Our proposed framework, InfuseNet, alleviates the magnification artefacts via infusing motion context from FrameFlow. Given onset, apex and pseudo-apex frames, we extract the magnified latent features as the input to the FrameMag network. Meanwhile, for FrameFlow, we extract the optical flow and strain features from the onset to apex frames. Following that, we perform successive layer feature infusion from FrameFlow to FrameMag, and then FrameMag predicts the eventual one-hot AUs of the samples.}
    \vspace{-4.0em}
    \label{fig:framework}
\end{figure*}

% % % \vspace{-0.5em}
\label{sec:methodology}
In this section, we introduce the InfuseNet framework which leverages the successive layers' features learnt with optical flow images to shred down the artefacts, as demonstrated in Figure \ref{fig:framework}. To formalize our methods, we name the network with magnified motion as \textbf{FrameMag} whereas the network with optical flow images as \textbf{FrameFlow}. For the process of InfuseNet, FrameMag acts as the main AU Detector whereas FrameFlow infuses its learnable context into the FrameMag model.

% \begin{figure*}[!hb]
%     \centering
%     \includegraphics[width=0.90\linewidth]{images/framework.png}
%     \caption{Our proposed framework, \textbf{InfuseNet} alleviates the Magnification Artefacts via infusing motion context from ResFlow. We select ResNet-18 as our backbone network for both feature inputs. Given onset, apex and pseudo-apex frames, we extract the magnified latent features as the input to the ResMag network. Whereas for ResFlow, we extract the optical flow and strain features from the onset to apex frames. Following that, we perform successive layer feature infusion from ResFlow to ResMag, and then ResMag predicts the eventual one-hot AUs of the samples.}
%     \label{fig:framework}
% \end{figure*}

% TAlk about magnification and flaws
\subsection{Model Inputs}
The inputs to InfuseNet are the optical flow image (motion context) and magnified latent features. Both of the inputs are computed between the motion of onset and apex frames. On a side note, the annotation of apex frames involves some uncertainties because micro-expressions occur in a fraction of a second, making the apex annotation subjective. Hence, in each iteration, we sample a pseudo-apex frame within the range of [0, 5] frame from the annotated apex frame's position with a fixed seed. This sampling strategy regularizes the feature learning process.

\subsection{Motion Latent Magnification}
While motion magnification is useful for magnifying the subtle signals of facial micro motions, it creates artefacts that confuse the feature learning process. The artefacts are caused by the reconstruction of amplified motions (decoding), which inaccurately projects the learnt latent features onto the eventual amplified motion image. Based on the learnable video motion magnification framework \cite{oh2018learning}, we propose to detach the decoder while fine-tuning the encoder and latent layer. 

Concretely, we extract magnified motions of such pairs, (A) Onset Frame - Apex Frame (B) Onset Frame - Pseudo-apex Frame. In data annotation, the definition of apex may have a variance of a few frames, hence, adding pair(B) accounts for the uncertainty in analysis.

% We extract the Magnified Latent Features of such pairs, followed by concatenating the latent features as an input to the FrameMag Network.

Adapting the learnable magnification framework \cite{oh2018learning}, we calculate the motion difference for pair (A) and (B) followed by an amplification process in the manipulator module respectively. This is formalized in Equation \ref{eq:mag} and weight sharing is performed to minimize the number of learnable parameters and constrain the feature consistency between pair (A) and (B). 

\begin{equation}
    Mag(x_{1}, x_{2}) = M_{manipulator}(M_{enc}(x_{1}, x_{2}))
    \label{eq:mag}
\end{equation}

Following that, both the magnified latent embeddings are concatenated channel-wise as an input to FrameMag to learn the magnified motion signals, this is formalized in Equation \ref{eq:only_mag_motion}. Notation-wise, $Mag$ represents the magnification module, $M_{manipulator}$ represents the magnification manipulator, $M_{enc}$ represents the magnification encoder and $x_{1}$, $x_{2}$ and $x_{3}$ represents the onset, apex and pseudo-apex respectively.

\begin{equation}
    feat(x_{1}, x_{2}, x_{3}) = concat(Mag(x_{1}, x_{2}), Mag(x_{1}, x_{3}))
    \label{eq:only_mag_motion}
\end{equation}

% Adapting the learnable magnification framework \cite{oh2018learning}, we use two sequential frames to calculate the motion difference followed by an amplification process in the manipulator module and reconstruct the amplified motion with a decoder. As micro facial motion can be obscure, decoding the amplified motion can introduce unwanted artefacts. Therefore, we propose extracting the outputs of the magnification manipulator instead of getting the decoded output, as shown in Equation \ref{eq:mag} and \ref{eq:only_mag_motion}. Notation-wise, $Mag$ represents the magnification module, $M_{manipulator}$ represents the magnification manipulator, $M_{enc}$ represents the magnification encoder and $x_{1}$, $x_{2}$ and $x_{3}$ represents the onset, apex and pseudo-apex respectively. $feat$ represents the feature extraction process that concatenates two magnified motions as an input to FrameMag.
% % \vspace{-1em}

% % \vspace{-1em}

% % \vspace{-1em}

\subsection{Artefacts Removal with Flow Infusion}
\label{sec:feat_infuse}

Besides the decoding problem that creates artefacts, as mentioned in Section \ref{sec:intro}, the other origin of motion artefacts comes from magnifying the subtle noise in the original frames. Thus, a guiding mechanism is needed to direct the feature learning process, ensuring emphasis on regions exhibiting plausible facial motions.

\subsubsection{Optical Flow Computation}
% todo move optical flow here
Optical flow and strain are the first and second derivatives of motion displacement respectively, which serve as decent candidates in guiding the feature learning process on "where to look". For that, we utilize \cite{sun2010secrets} and \cite{shreve2009towards} to compute the motion changes from onset to apex frames. Equation \ref{eq:optical_flow} shows the computation of horizontal and vertical motion changes whereas Equation \ref{eq:strain_short} shows the concise computation of optical strain.
% % \vspace{-0.5em}
\begin{equation}
	\vec{v} = [p = \frac{dx}{dt}, q = \frac{dy}{dt}] ^\mathrm{T}
    \label{eq:optical_flow}
\end{equation}
where $dx$ and $dy$ represent the estimated changes in pixels along the $x$ and $y$ dimension respectively while d\textit{t} represent the change in time,

% % \vspace{-1em}
\begin{equation}
	\epsilon = \frac{1}{2}[\nabla u + (\nabla u)^T]
    \label{eq:strain_short}
\end{equation}
where it calculates the second-order derivatives of horizontal and vertical displacements.

To fully utilize the strength of these two features, we group them as a feature cube known as an optical flow image (optical flow + optical strain) as formalized in Equation \ref{eq:optical_flow_image}.
% % \vspace{-0.5em}
\begin{equation}
    I_{OF} = ({f_x, f_y, \epsilon})
    \label{eq:optical_flow_image}
\end{equation}
where $I_{OF}$ is the optical flow image, $f_x$ and $f_y$ are the flow motion along the horizontal and vertical directions respectively, and $\epsilon$ is the optical strain as computed in Equation \ref{eq:strain_short}.

% Focus on improving this part
% We mention how it is done here. but the WHYs not so much
% re-reference the whole framework here. describe the framework here
\subsubsection{Feature Infusion}
In designing the fusion task, late fusion could be insufficient because the motion artefacts are affecting FrameMag to learn the genuine motion features. As circumvention, fusion should be performed at each layer of the backbone model to ensure that the artefacts are removed before passing it for feature learning. Hence, we propose a unitary successive layer-wise fusion from FrameFlow into the FrameMag network as demonstrated in Figure \ref{fig:framework}.

Referring to Figure \ref{fig:framework}, after extracting magnified latent features and optical flow image, we perform feature learning in two separate, identical backbone models. In each network block, the learnt optical flow features are infused into their counterparts with magnification inputs, thereby removing the artefacts before passing them to subsequent layers. Concretely, as demonstrated in Equation \ref{eq:opticalflow_feat}, the features learnt with optical flow image are normalized to act as a soft attention mask, indicating the region of interest where motion exists. 

\begin{equation}
    A^{l_n}_{of} = Norm(F^{l_n}_{of})
    \label{eq:opticalflow_feat}
\end{equation}
where $A$ is the min-max normalized features from the FrameFlow module.

Then, the normalized features are infused layer-wise successively via Hadamard Product into the FrameMag network block as demonstrated in Equation \ref{eq:infusion}. 

\begin{equation}
    F^{l_n}_{infusion} = F^{l_n}_{feature} \odot A^{l_n}_{of}
    \label{eq:infusion}
\end{equation}
where $F^{l_n}_{feature}$ represents the feature map from FrameMag of $layer_n$ and $F^{l_n}_{infusion}$ represents the infused feature map where magnification artefacts are reduced. 

The infused features are passed to the subsequent network block as shown in Equation \ref{eq:next_layer}.

\begin{equation}
    F^{l_{n+1}}_{feature} = FrameBlock^{l_{n+1}} ( F^{l_n}_{infusion} )
    \label{eq:next_layer}
\end{equation}
where $FrameBlock^{l_{n+1}}$ shows that the subsequent $FrameBlock$ receives the infused features computed from previous block.

Via the Hadamard product, the face areas without optical motions are filtered off thereby, reducing the artefacts. This ensures the patches retained for feature learning are both \textit{motion-active} and \textit{magnified} with minimal magnification artefacts remaining.

% % \vspace{-0.5em}
\section{Experiments \& Discussion}
We detail the benchmarking process and results, followed by ablation studies in the following subsections.
% % \vspace{-0.5em}
\subsection{Dataset \& Protocol}
The datasets that are utilized in our experiments are CASME \cite{yan2013casme}, CASME II \cite{yan2014casme}, CAS(ME)$^{3}$ \cite{li2022cas}, 4DME \cite{li2022deep}, MMEW \cite{ben2021video} and SAMM \cite{davison2016samm}. As proposed by \cite{varanka2023data}, we combine these six databases with overlapping AUs, as the definition of emotional labels can vary across different databases. We use the protocol known as Composite Database 6 Micro-Expressions (CD6ME) \cite{varanka2023data} that performs a leave-one-database-out cross-validation. In detail, one database will be opted out as a test set while the rest will be used for training, the process is repeated for a different database until all databases have been tested once which is followed by averaging their Macro-F1 scores. This protocol is chosen for its ability to unify multiple databases, enabling a fairer comparison of AU detection.

\subsection{Experimental Setup}
The experiments were carried out with an Adam optimizer with a learning rate 0.001 and exponential learning rate decay (ExponentialLR) with a gamma of 0.9. Both FrameMag and FrameFlow are fine-tuned with cross-entropy loss without weight sharing for 50 epochs. Both networks are pretrained from ImageNet. The magnification factor used in InfuseNet is 10.

% % % \vspace{-1em}

\subsection{Comparison with State-of-the-art}
We benchmark our proposed methods in Table \ref{tab:benchmarking_table} for each AU class and the average Macro-F1 score. When comparing different ResNet backbones, we observe a decline in performance as model depth increases (e.g., ResNet-34, ResNet-50), indicating that deeper networks may suffer from increased overfitting. On the other hand, ResNet-10 likely underfits due to its limited depth. These findings suggest that (1) ResNet-18 offers the best balance, making it the most optimal choice for further experimentation. (2) Using deeper networks is unrelated to improved performance.

Based on Table \ref{tab:benchmarking_table}, we can observe that InfuseNet(Res18) improves by 4\% compared to the backbone without infusement and improves by 1.8\% compared to the reported baseline, SSSNet under CD6ME protocol. In a brief review of performance across AU classes, InfuseNet(Res18) improves 5/7 smaller AU classes ( distribution $<$ 10\% ) and retains 4/5 bigger AU classes ( distribution $>$ 10\% ) compared to ResNet18. This demonstrates that the infusion process enables increased coverage of authentic signals while diminishing the motion artefacts. 

Besides that, we also compare the efficacy between InfuseNet(ViT-Tiny, patch-based learning) and InfuseNet(Res18, receptive-based learning). Based on Table \ref{tab:benchmarking_table}, we have discovered that InfuseNet(ViT-Tiny) improvement over its backbone without infusement is lower than that compared to InfuseNet(Res18). We deduce that the local receptive field in CNN reduces the toxicity of magnification artefacts. In contrast, the global receptive field of Vision Transformer increases the inclusion of artefacts, which can be explained by comparing ViT-Tiny and ResNet18 fine-tuned with magnification inputs. Therefore, the limited gain weakly correlates with stronger backbones, as different architectures could interact with the magnification mechanism in varying ways.

Compared to a closely related approach that employs adaptive magnification (LED)\cite{varanka2023learnable}, InfuseNet (with ResNet18) achieves an improvement of over 14\%. This indicates that InfuseNet enhances the adaptiveness of magnification more effectively than LED, which relies solely on learning Eulerian dynamics.

Apart from improving the general performance, we can observe that our proposed method with feature infusion has brought improvements to AU classes with minimal samples while retaining a certain extent of performance in bigger classes compared to the baseline SSSNet \cite{varanka2023data} results (State of the art in CD6ME protocol).

\begin{table}[!ht]
    \centering
    \vspace{-2em}
    \caption{Benchmarking table compares the performance of the proposed algorithm with baseline result, with the percentage of samples across AUs listed.}     
    \footnotesize
    \scalebox{0.62}{
    \begin{tabular}{|c|c|c|c|c|c|c|c|c|c|c|c|c|c|c|c|}
        \hline
        Method & Input Type & AU1 & AU2 & AU4 & AU5 & AU6 & AU7 & AU9 & AU10 & AU12 & AU14 & AU15 & AU17 & Average\\
        & (OF, Mag) & (12\%) & (11\%) & (28\%) & (5\%) & (2\%) & (10\%) & (5\%) & (3\%) & (7\%) & (11\%) & (2\%) & (3\%) & \\   
        \hline
        LBP-TOP \cite{varanka2023data} & OF & 0.4160 & 0.3670 & 0.6200 & 0.0000 & 0.0000 & 0.0000 & 0.0170 & 0.0000 & 0.0000 & 0.0350 & 0.0000 & 0.0000 & 0.1210 \\   

        Off-ApexNet \cite{gan2019off} & OF & \textbf{0.7490} & 0.7020 & 0.8630 & 0.1350 & 0.0330 & 0.4450 & \textbf{0.3660} & 0.1830 & 0.3200 & 0.3770 & 0.1900 & 0.3820 & 0.3950 \\
        
        SSSNet \cite{varanka2023data} & OF & 0.7460 & 0.7210 & \textbf{0.8750} & 0.1360 & 0.0530 & \textbf{0.4860} & 0.2030 & 0.1930 & 0.3620 & \textbf{0.4080} & 0.2270 & \textbf{0.4470} & 0.4050 \\     

        ResNet10 \cite{varanka2023data} & OF & 0.6870 & 0.6520 & 0.8380 & 0.0830 & 0.0640 & 0.3960 & 0.1100 & 0.0680 & 0.3100 & 0.2960 & 0.0880 & 0.3960 & 0.3320 \\
        
        ResNet18 \cite{varanka2023data} & OF & 0.6018 & 0.6541 & 0.6595 & 0.1612 & 0.1645 & 0.4362 & 0.2227 & 0.2378 & 0.3828 & 0.3298 & 0.3152 & 0.4276 & 0.3828 \\

        ResNet18 \cite{varanka2023data} & Mag & 0.4739 & 0.4252 & 0.6276 & 0.1685 & 0.1321 & 0.3442 & 0.2291 & 0.1576 & 0.3230 & 0.3788 & 0.2338 & 0.3979 & 0.3243 \\

        ResNet34 \cite{varanka2023data} & OF & 0.7210 & 0.7170 & 0.8580 & 0.1120 & 0.0540 & 0.3870 & 0.1600 & 0.1350 & 0.3650 & 0.3990 & 0.1480 & 0.3920 & 0.3710 \\

        ResNet50 & OF & 0.6474 & 0.6841 & 0.7525 & 0.1015 & 0.0228 & 0.3825 & 0.1561 & 0.0476 & 0.2970 & 0.2476 & 0.2160 & 0.2877 & 0.3202 \\

        LED \cite{varanka2023learnable} & Mag & 0.5270 & 0.4570 & 0.6370 & 0.0790 & 0.0070 & 0.1930 & 0.1360 & 0.0850 & 0.2650 & 0.3670 & \textbf{0.3300} & 0.3170 & 0.2830 \\

        ViT-Tiny \cite{wu2022tinyvit} & OF & 0.6924 & \textbf{0.7353} & 0.7984 & \textbf{0.2129} & 0.1510 & 0.4785 & 0.2371 & 0.1987 & 0.3742 & 0.3877 & 0.2871 & 0.3999 & 0.4128 \\

        ViT-Tiny \cite{wu2022tinyvit} & Mag & 0.4942 & 0.4896 & 0.6469 & 0.1389 & 0.1620 & 0.3296 & 0.1941 & 0.1259 & 0.3041 & 0.3196 & 0.1723 & 0.3281 & 0.3088 \\

        ResNet18(Late-fusion) & OF, Mag & 0.4820 & 0.4341 & 0.5962 & 0.1588 & 0.1454 & 0.3157 & 0.2161 & 0.1179 & 0.3020 & 0.3057 & 0.1467 & 0.3033 & 0.2937 \\        

        \hline
        InfuseNet(ViT-Tiny) & OF, Mag &
        0.6951 & 0.7100 & 0.7743 & 0.1917 & \textbf{0.1734} & 0.4361 & 0.3227 & 0.2287 & 0.3892 & 0.3513 & 0.2902 & 0.4431 & 0.4172 \\        
        
        InfuseNet(Res18) & OF, Mag &
        0.6997 & 0.6804 & 0.7921 & 0.1929 & 0.1572 & 0.4438 & 0.3263 & \textbf{0.2987} & \textbf{0.3987} & 0.3523 & 0.2973 & 0.4364 & \textbf{0.4230} \\
        
        \hline 
    \end{tabular}
    }  
    \vspace{-2em}
    % \caption{OF stands for optical flow, whereas Mag stands for magnified features.}
    \label{tab:benchmarking_table}
\end{table}

\begin{figure}
    \centering
    % % \vspace{-0.3em}
    \includegraphics[width=1.0\linewidth]{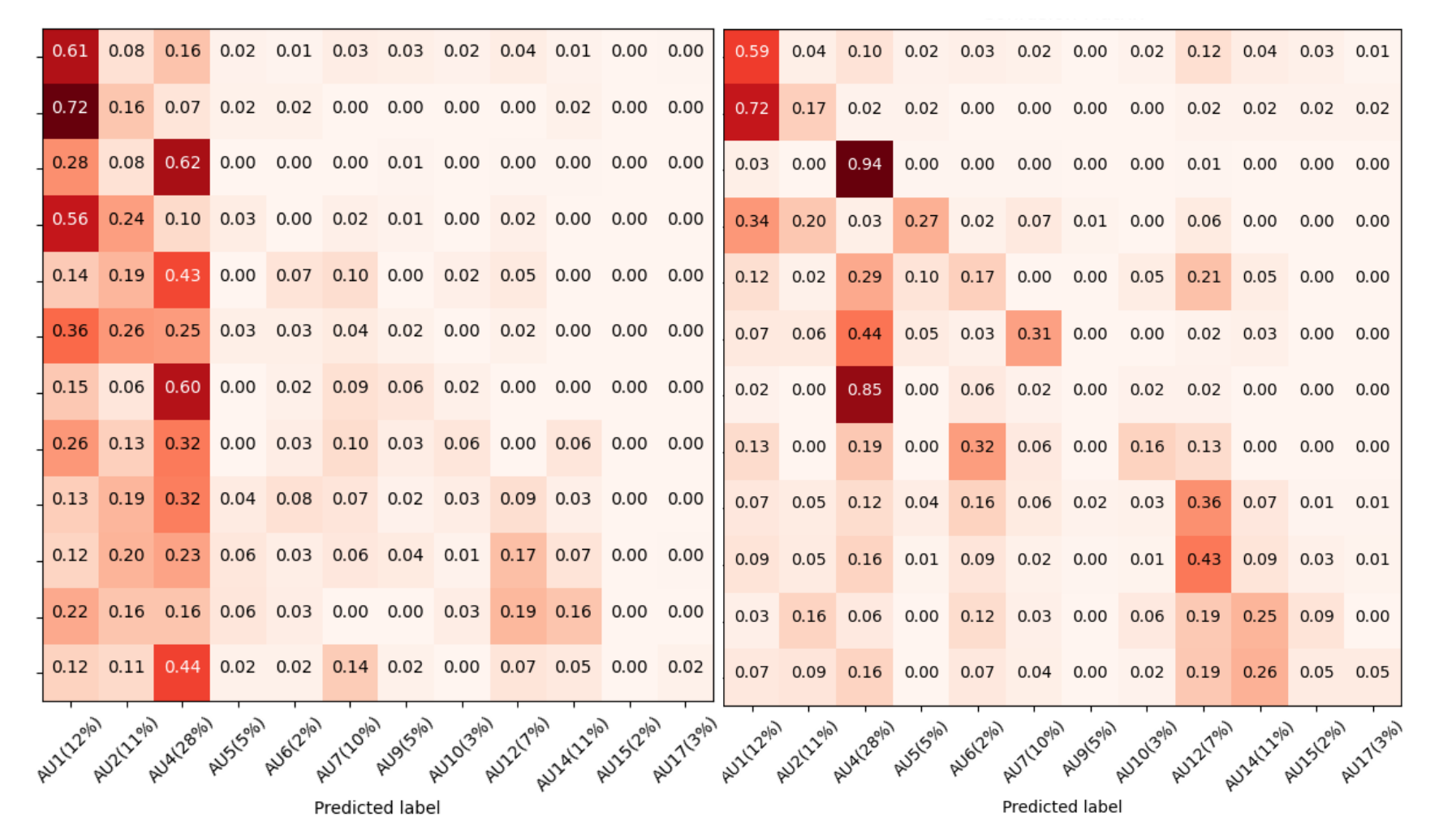}
    % % \vspace{-2em}
    \caption{The confusion matrix of SSSNet(Left) and InfuseNet-Res18(Right). In comparison, InfuseNet-Res18 yields a more balanced performance across AU classes as compared to SSSNet.}
    \vspace{-2.0em}
    \label{fig:confusion_matrix}
\end{figure}

Furthermore, we compare our proposed InfuseNet(Res18) with the baseline method SSSNet (SSSNet is the state-of-the-art results reported in CD6ME protocol proposal \cite{varanka2023data}) in Figure \ref{fig:confusion_matrix}. We can observe that the balanced performance across classes has been slightly improved instead of saturating over the bigger classes such as AU 1, 2 and 4. In addition, we also have discovered that InfuseNet(Res18) improves balance performance between the upper(AU1-9) and lower(AU10-17) face region AUs, demonstrating localization effects despite it being beyond the focus of our current work.

\subsection{Ablation Study}

\subsubsection{Infusion vs No Infusion}
To determine the efficacy of feature infusion, we expand the ablation study to include individual methods, late fusion via feature concatenation and InfuseNet as demonstrated in Table \ref{tab:fusion_vs_nofusion}. Based on Table \ref{tab:fusion_vs_nofusion}, we observe that InfuseNet surpasses both individual models and their late fusion. We deduce that successive layer infusion eliminates the artefacts from being learnt, contrary to late fusion where the artefacts remain throughout the feature learning process.

% individually, FrameMag performs 10\% lower than that in FrameFlow and late fusion of their features worsens the performance too as the magnification artefacts may have steered both the methods to learn the noises instead of the authentic signals. With InfuseNet, we can observe a significant improvement, as the successive layer infusion from FrameFlow eliminates the artefacts that hinder the feature learning process.

% To determine the efficacy of feature infusion, we compare models with and without infusion, which are InfuseNet and FrameMag. Based on the results in Table \ref{tab:fusion_vs_nofusion}, we have found a large gap in performance difference between the non-infusion and infusion techniques. This discrepancy demonstrates that the optical flow image can provide a soft attention mask as a context to FrameMag, thus eliminating the image artefacts generated during the magnification process.

% \begin{table}[!ht]
%     \centering 
%     \caption{The comparison between no infusion and infusion.}    
%     \begin{tabular}{|c|c|}
%         \hline    
%         Model & Average MF1\\
%         \hline
%         FrameMag & 0.2877 \\
%         InfuseNet & \textbf{0.4230} \\
%         \hline
%     \end{tabular}

%     \label{tab:fusion_vs_nofusion}
% \end{table}

\begin{table}[!ht]
    \centering 
    % % % \vspace{-1.5em}
    \caption{Ablation study for the infusion module.} 
    % % % \vspace{-0.5em}
    % \scalebox{}{ % Scale table by 80%
    \begin{tabular}{|c|c|c|c|c|c|}
        \hline    
        Model & OF Images & Magnification & Infusion & Late-fusion & Avg. F1\\
        \hline
        FrameFlow & \checkmark  &  &  &  & 0.3828 \\
        FrameMag &  & \checkmark &  &  & 0.2877  \\
        FrameFlow + FrameMag & \checkmark & \checkmark &  & \checkmark & 0.2937 \\
        InfuseNet & \checkmark & \checkmark & \checkmark & &  \textbf{0.4230} \\
        \hline
    \end{tabular}
    % }
    % % \vspace{-0.5em}
    \label{tab:fusion_vs_nofusion}
\end{table}

\subsubsection{Latent Magnified Features vs Decoded Magnified Image}
To test the efficacy of latent magnified features, we compare the results between the use of latent magnified features and decoded magnified images, as shown in Table \ref{tab:latent_vs_decoded}. The results signify that using the latent magnified features can perform 5\% better than the decoded magnified image, proving that our hypothesis of using latent magnified features can further eliminate the magnified artefacts.

% % \vspace{-1em}
\begin{table}[!ht]
    \centering
    \caption{The comparison between using latent magnified features and decoded magnified image as model input to FrameMag.}
    % % \vspace{-1em}
    \scalebox{0.9}{
    \begin{tabular}{|c|c|}
        \hline
         Input to Model & Average MF1 \\
         \hline
         Decoded Magnified Image & 0.3758 \\
         Latent Magnified Features & \textbf{0.4230}  \\
         \hline
    \end{tabular}
    }
    \label{tab:latent_vs_decoded}
\end{table}

\vspace{-1em}
\subsubsection{Magnification Factor}
\begin{figure}
    \centering
    \includegraphics[width=0.85\linewidth]{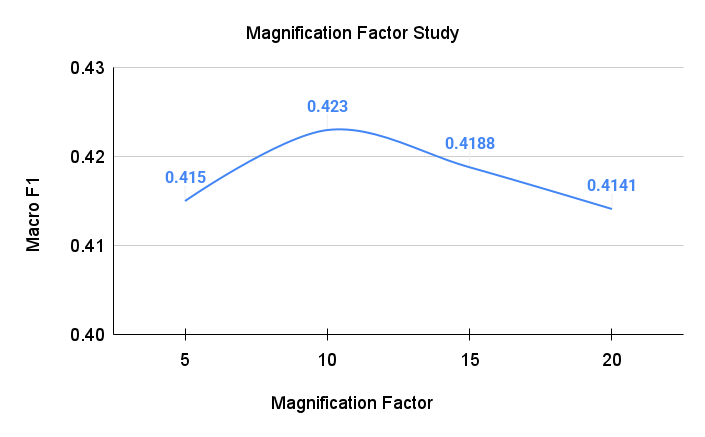}
    % % \vspace{-1em}
    \caption{The comparison of magnification factors. This side experiment identifies the optimal magnification factor for our proposed framework.}
    % % \vspace{-1em}
    \label{fig:magFactors_benchmark}
\end{figure}

To select a suitable magnification factor for \textbf{InfuseNet}, we perform an ablation study by alternating the magnification factor from 5 to 20 in the range of 5 incrementally to test its sensitivity. Based on Figure \ref{fig:magFactors_benchmark}, we observe that a magnification factor of 10 yields the most optimum result while exceeding this factor worsens the general performance. We deduce that magnification artefacts build up as the magnification factor increases, eventually distorting the motion signals post-infusion. Nevertheless, the difference between the optimum magnification factor and the worst-performing magnification factor is a mere 0.0089, demonstrating the capability of infusion in alleviating motion artefacts in the scenario where the level of motion artefacts is twofold of the optimum magnification factor.

% need a table here

% % \vspace{-0.5em}

\subsection{AU Visualization}
\begin{figure}
    \centering
    % % % \vspace{-0.5em}
    \includegraphics[width=1.0\linewidth]{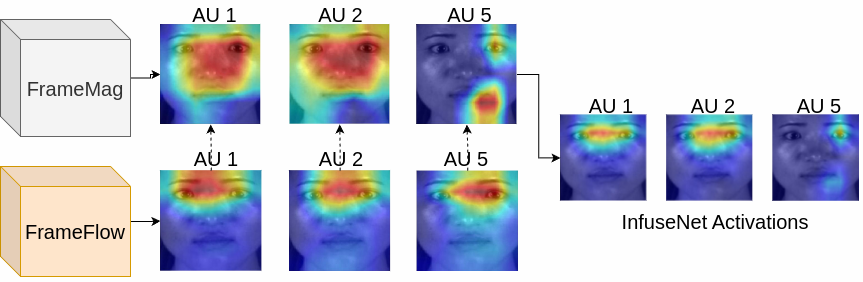} % pdf here will crash latex
    % % \vspace{-2em}
    \caption{Visualization of the activations to simulate how infusion can eliminate the artefacts from FrameMag. In this sample, we can observe that the infusion helps to pinpoint the precise AU location without sacrificing the features from magnification.}
    % % \vspace{-2em}
    \label{fig:grad-cam}
\end{figure}

\vspace{-0.5em}
To strengthen our understanding of the infusion process, we visualize the activations of our network with Grad-CAM \cite{jacobgilpytorchcam} \cite{selvaraju2017grad} to determine how close the activations are to the AU motions and the efficacy of the infusion process in removing artefacts from magnification features. As shown in Figure \ref{fig:grad-cam}, we can observe that FrameFlow contains a relatively smaller feature map in contrast to FrameMag. This shows that magnified features can enclose higher motion activities which are barely observable in optical flow features. However, magnification can induce artefacts as shown in AU5, where both the eyes and mouth region in FrameMag are activated, whilst supposedly an upper lid raiser. With InfuseNet, it eliminates the redundant activations triggered by magnification artefacts, which eventually shifts the focus of the feature learning process to the micro-motion region.

\vspace{-3em}

% \subsubsection{Feature Infusion Direction}
% In fusing optical flow images and magnified motions, hypothetically, using magnified motion as a soft attention mask to optical flow images makes as much sense as vice versa, hence, we opt to investigate if there is a difference in the direction of feature infusion. As such, we compare the infusion direction as shown in Table \ref{tab:fusion_direction}. Based on the results, we observe that weighing FrameMag with optical flow images works better than vice versa. This experiment shows that magnified motions can work better given that a decent attention mask is provided to filter off the artefacts. 

% \begin{table}[!ht]
%     \centering
%     \caption{Fusion comparison by switching the Main and Auxiliary Branches.}    
%     \begin{tabular}{|c|c|c|}
%         \hline
%          Fusion Main Branch & Auxiliary & Average MF1 \\
%          \hline
%          FrameFlow & FrameMag & 0.3904 \\
%          FrameMag & FrameFlow & 0.4230 \\
%          \hline
%     \end{tabular}

%     \label{tab:fusion_direction}
% \end{table}

% \input{tex/discussion}
\section{Conclusion \& Future Work}
\label{sec:conclusion}
In this work, we address the issue of alleviating the magnification artefacts, thereby enhancing the efficacy of motion magnification in micro-expression. We design the successive layer features of the optical flow image to eliminate magnification artefacts and minimize their generation by directly using the magnified latent features. Through the feature infusion process, the refined magnified motions significantly enhance the feature learning process for micro-expression AU detection, achieving performance that surpasses state-of-the-art results. Nevertheless, exploring magnification under noisy or high-variation lighting conditions is valuable, though it falls outside the scope of this study. Future work could focus on enhancing magnification robustness across diverse conditions and improving AU localization to achieve more accurate detection.

% Via the feature infusion process, the cleaner magnified motions greatly help the feature learning process in the micro-expression AU detection task and surpass the state-of-the-art results. 

% maybe can add it here from confusion matrix
% Interestingly, Figure \ref{fig:confusion_matrix} shows that there are two potential problems:
% \begin{itemize}
%     \item The imbalance sample distribution creates an imbalanced learning problem in which smaller classes are often misclassified as larger ones. For instance, we can observe that 85\% of AU9 has been misclassified as AU4.
%     \item There is a significant percentage of samples of AU14, AU15 and AU17 are misclassified as AU12 or AU14. These AUs are located around the mouth region of the subjects, hence we deduce that the current framework is struggling to differentiate the fine difference among these AUs. Similarly, for the eyes region, 72\% of AU2 and 34\% of AU5 are misclassified as AU1, further demonstrating the need for AU localization. 
% \end{itemize}

%
%
%

\section*{Acknowledgements}
This work was supported by the Research Council of Finland (former Academy of Finland) Academy Professor project EmotionAI (grants 336116, 345122, 359854), the University of Oulu \& Research Council of Finland Profi 7 (grant 352788), EU HORIZON-MSCA-SE-2022 project ACMod (grant 101130271), and Infotech Oulu. The authors wish to acknowledge CSC – IT Center for Science, Finland, for computational resources.

% \begin{credits}
% \subsubsection{\ackname} A bold run-in heading in small font size at the end of the paper is
% used for general acknowledgments, for example: This study was funded

% \subsubsection{\discintname}
% It is now necessary to declare any competing interests or to specifically
% state that the authors have no competing interests. Please place the
% statement with a bold run-in heading in small font size beneath the
% (optional) acknowledgments\footnote{If EquinOCS, our proceedings submission
% system, is used, then the disclaimer can be provided directly in the system.},
% for example: The authors have no competing interests to declare that are
% relevant to the content of this article. Or: Author A has received research
% grants from Company W. Author B has received a speaker honorarium from
% Company X and owns stock in Company Y. Author C is a member of committee Z.
% \end{credits}
%\clearpage
%\newpage
\bibliographystyle{splncs04}
\bibliography{egbib}
% \bibliography{egbib}
% \end{thebibliography}
\end{document}